\documentclass{article}
\usepackage{spconf, graphicx}
\usepackage{booktabs}
\usepackage{multirow}
\usepackage{amsmath}
\usepackage{amssymb}
\usepackage{caption}
\usepackage{subcaption}
\usepackage{cite}
\usepackage{wrapfig}
\usepackage{comment}
\usepackage{hyperref}
\usepackage{nicefrac}       
\usepackage{adjustbox}
\usepackage{makecell}
\usepackage{multirow}
\usepackage{array}
\usepackage{amsthm}
\usepackage{stmaryrd}
\usepackage{yfonts}
\usepackage{tabularx}
\usepackage[T1]{fontenc}

\newcommand{\SthreeDW}{\textit{\textbf{S}yn\textbf{3DW}ound}}

\title{Syn3DWound: A Synthetic Dataset for 3D Wound Bed Analysis}

\name{
\begin{tabular}{c}
L\'eo Lebrat$^{1,3}$, Rodrigo Santa Cruz$^{1,3}$, Remi Chierchia$^{1,3}$, Yulia Arzhaeva$^{1}$, \\ \textit{Mohammad Ali Armin$^{1}$, Joshua Goldsmith$^{1}$, Jeremy Oorloff$^{1}$, Prithvi Reddy$^{1}$, Chuong Nguyen$^{1}$}, \\ 
\textit{Lars Petersson$^{1}$, Michelle Barakat-Johnson$^{2}$, Georgina Luscombe$^{2}$, Clinton Fookes$^{3}$}, \\  
\textit{Olivier Salvado$^{1}$, David Ahmedt-Aristizabal$^{1,3}$ }
\end{tabular}
}
\address{
$^{1}$ Imaging and Computer Vision Group, CSIRO Data61, Australia \\
$^{2}$ Faculty of Medicine and Health, University of Sydney, Australia \\
$^{3}$ SAIVT, Queensland University of Technology, Australia \\
\tt\normalsize \{leo.lebrat,david.ahmedtaristizabal\}@data61.csiro.au\\
\normalsize Code and Dataset : \url{https://lebrat.github.io/Syn3DWound/} \vspace*{-.5cm}
}

\begin{document}
\ninept
\maketitle

\begin{abstract}
Wound management poses a significant challenge, particularly for bedridden patients and the elderly. Accurate diagnostic and healing monitoring can significantly benefit from modern image analysis, providing accurate and precise measurements of wounds. 
Despite several existing techniques, the shortage of expansive and diverse training datasets remains a significant obstacle to constructing machine learning-based frameworks. 
This paper introduces \SthreeDW, an open-source dataset of high-fidelity simulated wounds with 2D and 3D annotations. 
We propose baseline methods and a benchmarking framework for automated 3D morphometry analysis and 2D/3D wound segmentation.
\end{abstract}

\begin{keywords}
Wound documentation, 3D reconstruction, 2D/3D wound segmentation.
\vspace{-8pt}
\end{keywords}

\section{Introduction}
\vspace{-5pt}

Chronic wounds, a widespread issue affecting individuals of all ages, represent a silent epidemic. It was estimated in 2019 that the prevalence of chronic wounds of mixed etiologies was 2.21 per 1000 population\cite{MARTINENGO20198}. Wound management is a major issue for bedridden patients in hospitals and elderly residents in aged care facilities. 
Wound management is challenging, and there is no standardized patient-centric care model. Wound documentation is crucial and should encompass a range of details such as location, size, surrounding skin condition, presence of undermining and tunneling, exudate, odor, or pain levels. Automated wound analysis by a computer system would allow accurate and precise diagnosis and assessment of the wound type, and enable quantitative assessment during healing, which could span months. Automated wound characterization offers a key advantage by allowing remote monitoring, eliminating the necessity for frequent and expensive physical examinations by medical specialists.  

\begin{figure}[!t]
\centering
\includegraphics[width=0.86\linewidth]{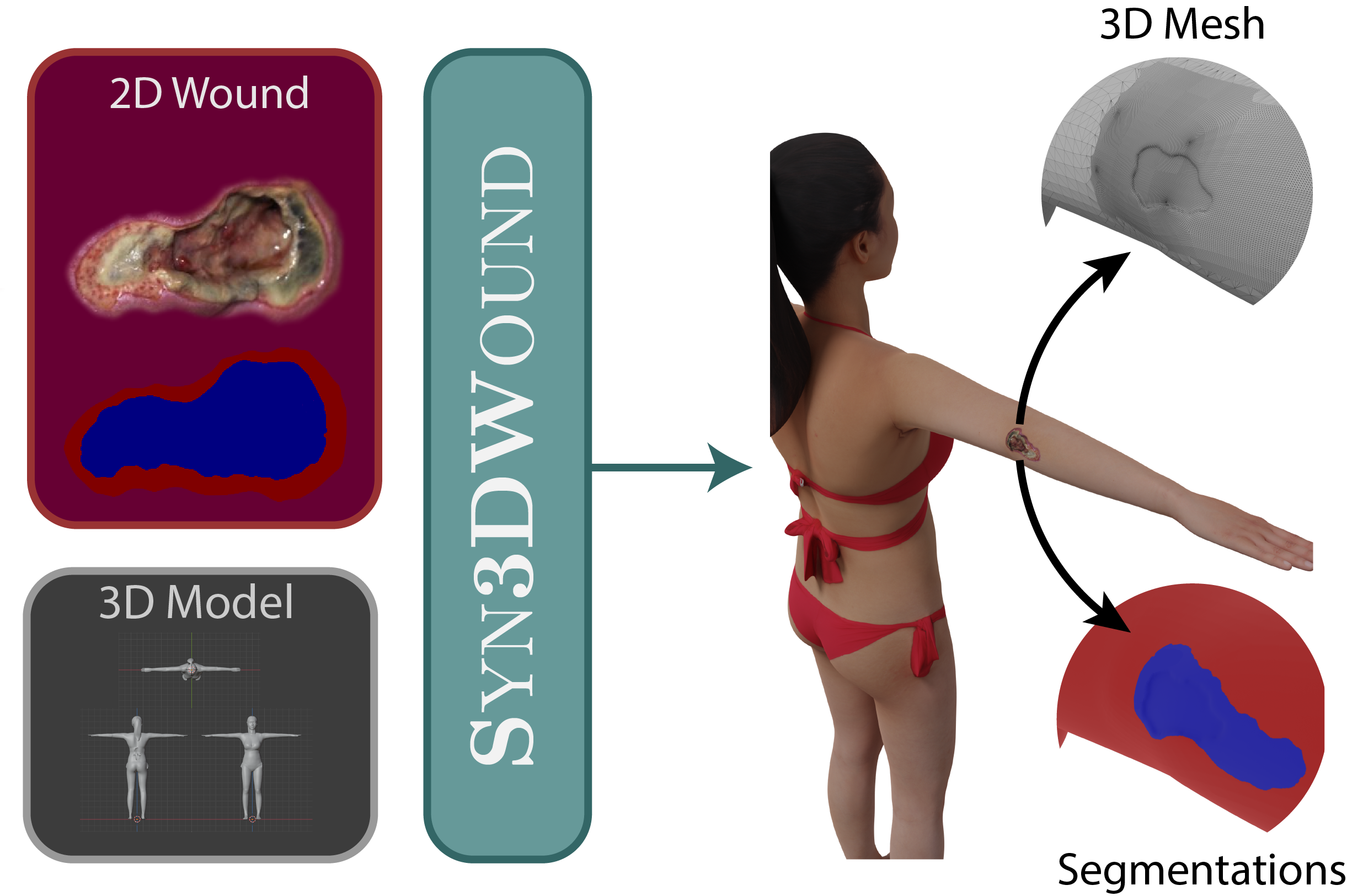}
\vspace{-3pt}
\caption{
\SthreeDW~aims to produce high-quality synthetic data with precise control of the environment and acquisition protocol from a 2D real-world wound and a 3D avatar. It allows the generation of extensive datasets for evaluating segmentation models. Furthermore, the camera's intrinsic and extrinsic are saved to analyze the performance of 3D reconstruction methods. 
}
\label{fig:wound_visualisation}
\vspace{-15pt}
\end{figure}

\begin{figure*}[!t]
\centering
\includegraphics[width=0.9\linewidth]{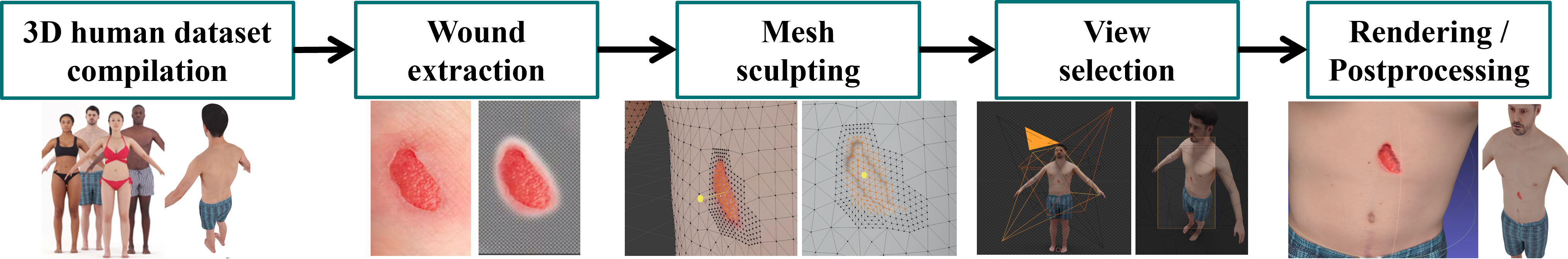}
\vspace{-3pt}
\caption{
Representations of the specific components involved in the synthetic wound generation. i) 3D human avatar, ii) Wound image and wound extraction, iii) Mesh sculpting including wound shape and placement in the human body, iv) View selection of the 3D human body avatar, and v) Rendering and postprocessing.
}
\label{fig:Pipeline_data_generation}
\vspace{-6pt}
\end{figure*}

Wound assessment based on photography/videos is challenging because of substantial variations in appearance and quality caused by different camera quality, lighting, and camera pose.
Data-driven vision-based technologies have been shown to improve wound assessment by enabling objective quantitative evidence for decision support~\cite{litjens2017survey}. Researchers have reported deep learning methods for 2D wound detection and classification~\cite{zhang2022survey}, wound segmentation~\cite{wang2020fully,oota2023wsnet,kendrick2022translating} or 2D wound image healing classification~\cite{oota2021healtech}.
However, 2D wound measurement techniques do not report wound depth, potentially overlooking a crucial aspect of the wound healing process. Additional challenges include identifying wound margins, variations in the wound's appearance due to changes in patient position, and the natural curvature of body parts such as the heel, toe, and lower leg.

Advanced 3D imaging technology, coupled with automated analysis methods, enables standardized and comprehensive image acquisition~\cite{ahmedt2023monitoring}. It could provide natural representation and measurements, especially for attributes that may be challenging to identify in 2D images~\cite{ahmedt2023monitoring,mirikharaji2023survey}. 
Automated wound analysis in 3D could assess the topology and textural features of wounds
~\cite{Liu2019WoundImages, MirzaalianDastjerdi2019MeasuringAlgorithms, Shirley2019AWounds, 
Barbosa2020AccurateMotion,
Sanchez-Jimenez2022SfM-3DULC:Area}, offering valuable clinical information.
%
A major bottleneck for training modern machine learning systems is obtaining high-quality training datasets and their associated ground truth (annotated by medical experts). Datasets that include 3D sensing are scarce, and collecting video of actual wounds is problematic: it has the potential to interfere with care, may include sensitive views, and can only be performed with limited camera and light setups. 
An alternative to collecting actual data is synthesizing images and their corresponding annotations, a strategy used in various domains, sometimes called digital twin~\cite{shamsolmoali2021image,mirikharaji2023survey}. Relevant to this paper,
Dai et al.~\cite{dai2021burn} generated textured burn wounds from a 3D human avatar as a synthetic annotated dataset. Sinha et al.~\cite{sinha2023dermsynth3d} used similar methods to create 2D images from 3D textured meshes with diverse skin tones and background scenes.

In contrast to existing methods, our proposed solution produces 2D synthetic data and precise 3D wound models, facilitating the evaluation of state-of-the-art 3D reconstruction methodologies (Fig.~\ref{fig:wound_visualisation}). This contribution is two-fold: Firstly, we introduce a 3D Wound synthetic dataset \SthreeDW, available for research purposes, with 2D and 3D ground truth. Secondly, we present baseline methods and evaluation protocols for i) 3D wound reconstruction, ii) 2D wound bed segmentation, and iii) 3D wound bed mapping, showcasing the merits of 3D wound analysis over 2D approaches.

\vspace{-8pt}
\section{Syn3DWound Dataset}

The synthetic views in \SthreeDW~ are generated using Blender, an open-source 3D computer graphic software, capable of producing realistic stills and videos by controlling the camera path. The user has the flexibility to manipulate wound characteristics, its location on the body, human body shape, and texture. The key steps are outlined in Fig.~\ref{fig:Pipeline_data_generation}.

The inputs consist of a 3D human body avatar, a 2D wound image, and a predefined 3D wound shape and location. Users can manually carve a wound onto the 3D human body avatar surface, specifying its depth and location. The visual appearance of the wound, along with its segmentation mask, is integrated into the avatar's texture files. 

The outputs include a 3D human body avatar featuring an attached wound, a collection of rendered images depicting various camera and environmental configurations, and all the necessary parameters for replicating the output. Beyond achieving pixel-perfect segmentation masks and comprehensive data generation, \SthreeDW~ also provides precise 3D models of the wound, essential for assessing the effectiveness of 3D methodologies.

We employed \textit{The Rendered people dataset}~\cite{Renderpeople_bundle_swimwear} and the \textit{3D Body Text dataset}~\cite{saint20183dbodytex}, which offer high-definition textured meshes of the human body in high resolution. For the 3D rendering engine, Cycles~\footnote{https://www.cycles-renderer.org/} was chosen for its enhanced light physics modelling and more lifelike rendering compared to routinely used real-time game graphics engines~\cite{sinha2023dermsynth3d}.

\begin{figure}[!ht]
\centering
\includegraphics[width=0.98\linewidth]{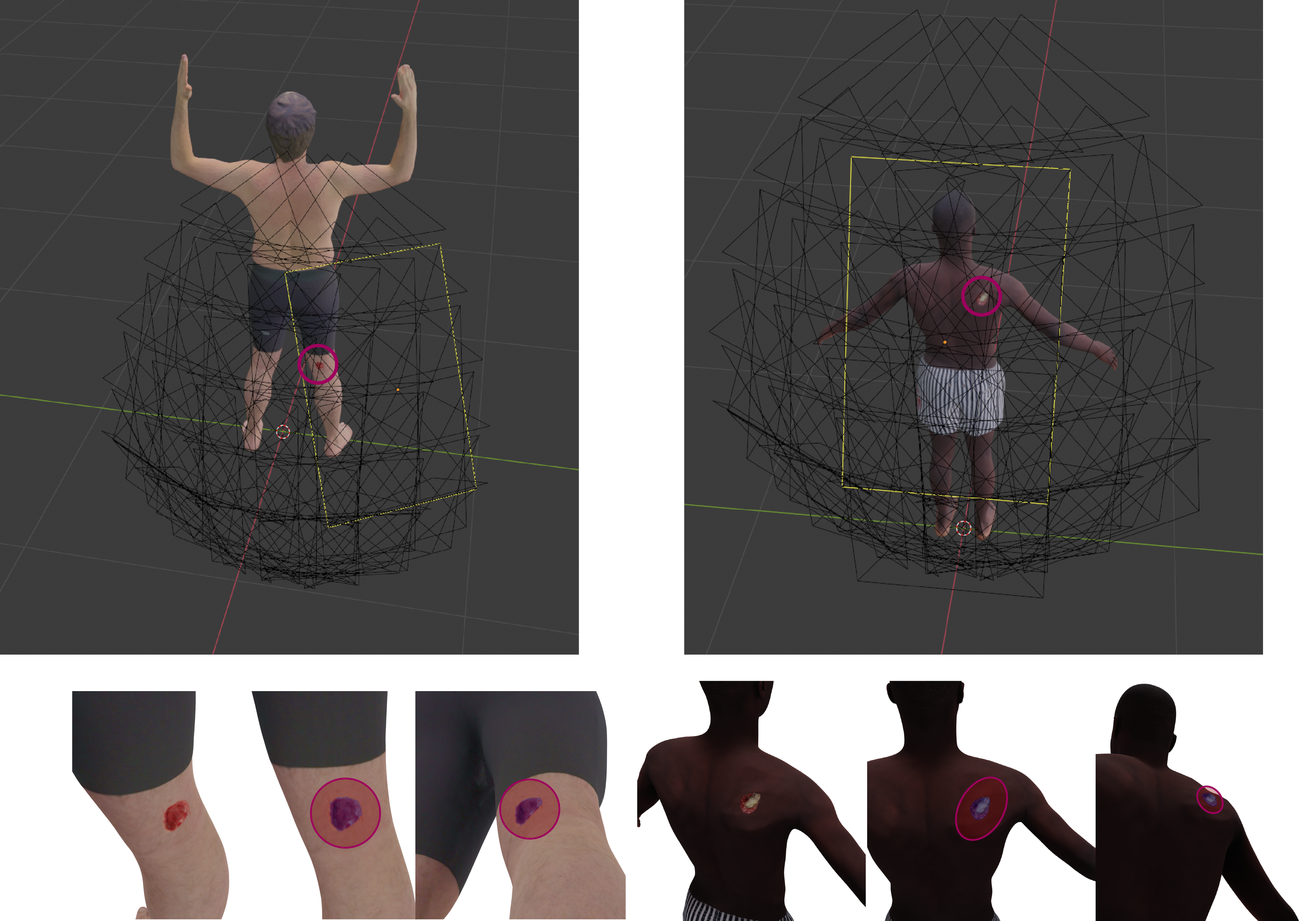}
\vspace{-3pt}
\caption{
Representations of 2D wound images and their corresponding segmentation maps are generated from various camera trajectories of the 3D wound models. The two models featured in this manuscript are a leg wound (left) and a shoulder wound (right).
}
\label{fig:visualisation_outputs}
\vspace{-9pt}
\end{figure}

After generating the 3D scene, users can create a camera path, allowing variations in the number of images for 3D reconstruction, as well as the ground truth for camera intrinsics and trajectory. For a particular wound, users can explore different observation angles,  camera resolutions, and lens characteristics, as depicted in the first row of ~Fig.~\ref{fig:visualisation_outputs}. 
To simulate imperfections present in real-world image acquisition, users can intentionally introduce either overexposure or apply motion/Gaussian blurring to the rendered images. 
Lighting aspects, such as the strength and the 3D placement of the light source, can also be adjusted at this stage, influencing the appearance of shadows in the rendered image. 


Ideally, wound characterization would include wound type, body location, size, variations in lighting conditions, and skin colour difference. Unfortunately, the availability of labelled data for 3D wound analysis has been limited. 
Existing datasets such as WoundSeg~\cite{oota2023wsnet}, DFUC2022~\cite{kendrick2022translating}, FUSeg Challenge~\cite{wang2021foot}, AZH wound care~\cite{wang2020fully}, and  Medetec~\cite{thomas2020medetec} primarily consist of 2D annotated images. 
WoundDB~\cite{krkecichwost2021chronic} provides stereo images with the potential for depth estimation investigations. However, these images are not sequential, which limits their utility for 3D wound reconstruction. In contrast, \SthreeDW~ provides perfect information, albeit simulated. Table~\ref{table:summary_datsets} compares \SthreeDW~ with these existing datasets.

\begin{table*}[!htb]
\caption{A comparison of the proposed 3D Wound bed dataset and the existing wound datasets.}
\vspace{-6pt}
\centering
\label{table:summary_datsets}
\resizebox{0.98\textwidth}{!}{%
\begin{tabular}{l c c c c l}
\toprule
\textbf{Dataset} & \textbf{Year} & \textbf{Modality} &  \textbf{Domain} & \textbf{Total Images} & \textbf{Wound Etiology} \\
\midrule
\textbf{Syn3DWound} (our) &  2023  & RGB & 2D/3D & 20 models $\dag$ & Pressure, trauma, arterial \\
WoundSeg~\cite{oota2023wsnet}      &  2023  & RGB   & 2D    & 2,686 & Diabetic, pressure, trauma, venous, surgical, arterial, cellulitis, and others \\
DFUC2022~\cite{kendrick2022translating}  &  2022  & RGB   & 2D    & 4,000 & Foot ulcer \\
WoundsDB~\cite{krkecichwost2021chronic}  &  2021  & RGB, Stereo, Thermal & 2D & 737 $\star$ & Venous ulcers, ischaemia, venous ulcers  \\
FUSeg Challenge~\cite{wang2021foot} &  2021  & RGB   & 2D    & 1,210 & Foot ulcer \\
AZH wound care~\cite{wang2020fully}  &  2020  & RGB   & 2D    & 1,109 & Foot ulcer \\
Medetec~\cite{thomas2020medetec}    &  NA    & RGB   & 2D    & 160   & Foot ulcer \\
\bottomrule
\multicolumn{6}{p{600pt}}
{
$\dag$ The \SthreeDW~dataset consists of 3D wound models that could generate diverse 2D wound images from different views of the same target wound.

$\star$  188 RGB, 188 thermal, 184 stereo, and 177 depth images are included in the WoundsDB database.
}
\end{tabular}}
\vspace{-12pt}
\end{table*}


\vspace{-6pt}
\section{Experiments and results}
In this section, we detail the evaluation protocol to perform 2D and 3D wound assessment of two 3D models, each representing a different ethnicity and depicted in Fig.~\ref{fig:visualisation_outputs}. Upon the acceptance of our paper, we will release a more extensive dataset, along with the code required to compute the evaluation metrics.

\vspace{-6pt}
\subsection{Baseline systems and evaluation metrics}


\begin{figure}[!ht]
\centering
\includegraphics[width=0.99\linewidth]{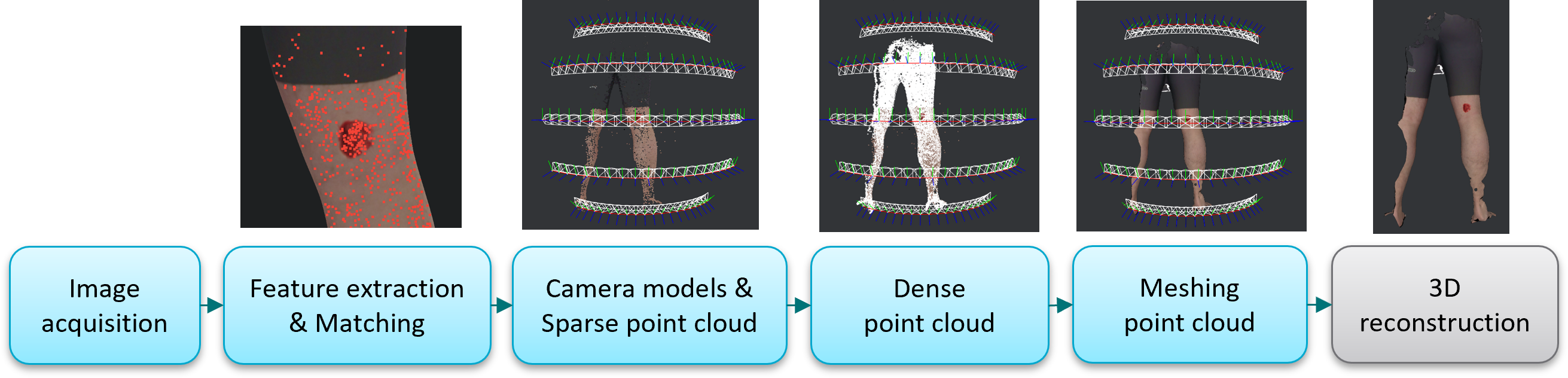}
\vspace{-3pt}
\caption{
Overview of a traditional framework for 3D reconstruction and analysis: sequential image collection, feature extraction and matching, camera models and sparse point cloud, dense point cloud, meshing point cloud, and 3D reconstruction.
}
\label{fig:3D_pipeline}
\vspace{-9pt}
\end{figure}

\noindent\textbf{3D wound reconstruction:}
A 3D reconstruction algorithm estimates the 3D geometry of an object from a collection of 2D images. The prevailing methods in the literature rely on standard projective geometry techniques such as structure-from-motion and multiview stereopsis~ \cite{Kumar2019AWounds,Shirley2019AWounds,Barbosa2020AccurateMotion}. However, new deep learning approaches for 3D scene rendering (\textit{e.g.} Neural Radiance Fields (NeRF)~\cite{Yu2022SDFStudio}), are becoming very competitive.
In this paper, we conduct a comparative analysis of two prominent open-source tools for 3D reconstruction: COLMAP~\cite{COLMAP} and Meshroom~\cite{Meshroom}. We also assess the performance of NeuS-Facto, a NeRF model tailored for surface extraction from the open-source SDFStudio toolbox~\cite{Yu2022SDFStudio}.

We compared the 3D reconstructed meshes with the ground-truth synthetic mesh, after alignment using three steps: i) align the camera positions of the ground-truth data with those estimated by the frameworks (by solving a Procrustes problem~\cite{golub2013matrix}); ii) crop both meshes using the ground-truth 3D mask for wound bed segmentation, followed by fine alignment using the Iterative Closest Point (ICP) algorithm (applied only to the cropped meshes); iii) apply the transformations to the original meshes, followed by cropping the wound area again to report performance on the wound area only. In Table~\ref{table:reconstruction_results}, we report the Average Symmetric Distance (ASD), Hausdorff Distance (HD90), and Normal Consistency (NC) metrics.

\begin{figure}[!t]
    \centering
    \includegraphics[width=.35\textwidth]{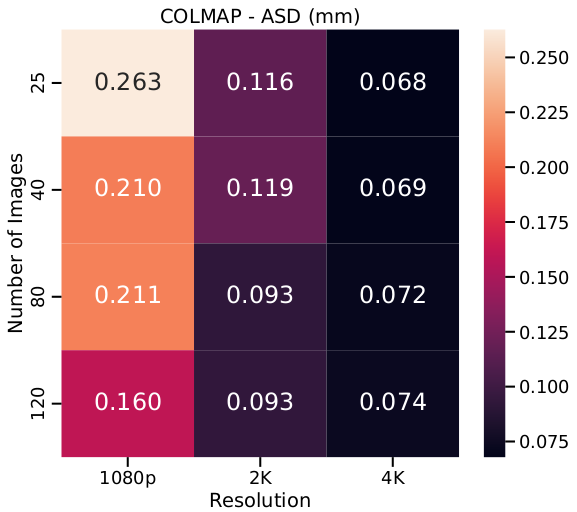}
    \vspace{-4pt}
    \caption{Evaluating 3D reconstruction outcomes across diverse image resolutions and quantities on the shoulder wound. We showcase results for the COLMAP pipeline using the ASD metric. However, similar trends are observed across different 3D reconstruction methodologies and evaluation metrics.}
    \label{fig:resolution_metrics}
    \vspace{-9pt}
\end{figure}

\begin{figure*}[!t]
    \centering
    \includegraphics[width=.94\textwidth]{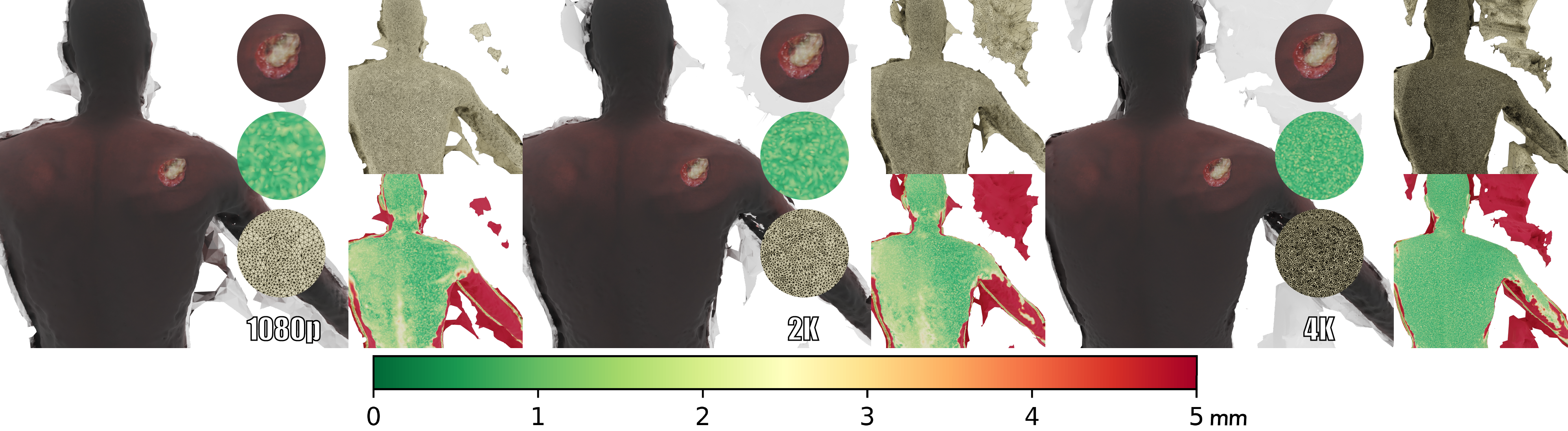}
    \vspace{-6pt}
    \caption{Assessing the impact of the image resolution (1080p to 4k) on mesh quality from Meshroom reconstruction on Sample 2 (shoulder wound). An increased camera resolution allows the reconstruction of more detailed geometries}
    \label{fig:importanceHR}
\vspace{-6pt}
\end{figure*}

The proposed pipeline facilitates benchmarking of 3D reconstruction methods and investigation into the influence of image features in the performance of the reconstruction method. 
Fig.~\ref{fig:resolution_metrics} shows the overall performance on the shoulder wound. COLMAP outperforms its competitor with increased image resolution. In every scenario, high-resolution images allow more fine-grained 3D reconstruction (see Fig.~\ref{fig:importanceHR}).

\vspace{4pt}
\noindent\textbf{2D wound segmentation:}
We trained a deep learning segmentation model SegFormer~\cite{xie2021segformer} on a dataset provided by DFUC2022~\cite{kendrick2022translating} and tested it on a set of images from \SthreeDW. From a predicted mask (A) and a ground truth mask (B), we compared the IoU score (Intersection over Union):$\frac{\lvert A \cap B \rvert}{\lvert A \cup B \rvert}$, and the Dice score: $\frac{2 \lvert A \cap B \rvert}{\lvert A \rvert + \lvert B \rvert}$.

\vspace{4pt}
\noindent\textbf{3D wound bed segmentation:}
We introduce a 3D wound segmentation technique that assigns binary labels to different regions of the reconstructed 3D models. We used a Meshroom-based texturing algorithm~\cite{griwodz2021alicevision} to project a set of 2D wound segmentation masks onto 3D mesh vertices labeled as background and wound bed.

Following the established standard ~\cite{taha2015metrics}, we report the Balanced Average Hausdorff distance (BAHD)~\cite{aydin2021usage}, defined as $\text{BAHD}( G, S ) = \frac{1}{2 |G| } \left(\mathcal{H}(G,S) + \mathcal{H}(S,G) \right)$, where $\mathcal{H}$ is the directed average Hausdorff distance and $|G|$ is the number of points in the ground truth wound segmentation. We also report recall $R=(T_p)/(T_p+F_n)$ and precision $P=T_p/(T_p + F_p)$, with $T_p$ the number of vertices from the 3D ground truth segmentation that are also in the 3D estimated segmentation, $F_p$ the number of vertices in the predicted segmentation that are missing from the ground truth segmentation, and $F_n$ is the number of the ground truth segmentation vertices missing from the predicted segmentation.


\begin{table}[!t]
\caption{Evaluating the performance of established 3D reconstruction pipelines by benchmarking the reconstruction of Sample 1 (leg wound), illustrated through a set of 300 2D images.}
\vspace{-6pt}
\centering
\label{table:reconstruction_results}
\resizebox{0.48\textwidth}{!}{%
\begin{tabular}{l c c c}
\toprule
\textbf{Methodology/Tool} &  $\downarrow$ \textbf{ASD}~\textit{(mm)} & $\downarrow$ \textbf{HD90}~\textit{(mm)} & $\uparrow$ \textbf{NC} \\
\midrule
Meshroom~\cite{griwodz2021alicevision}    & 0.394 & 1.052 & 0.933 \\
COLMAP~\cite{COLMAP}                      & {\bf 0.161} & {\bf 0.397} & 0.953 \\
NeuS-Facto (SDFStudio)~\cite{Yu2022SDFStudio} & 0.166 & 0.404 & {\bf 0.960} \\
\bottomrule
\end{tabular}}
\vspace{-8pt}
\end{table}

\vspace{-6pt}
\subsection{Results and discussion}

\noindent\textbf{Influence of the quality of the images:}
While a recent study explores the use of synthetic images for dermatological assessments~\cite{sinha2023dermsynth3d} with relatively small $512\times 512$ images, we propose adopting Cycles, a powerful rendering engine that outperforms Open3D's physic-based renderer or Unity3D~\footnote{https://docs.unity3d.com/ScriptReference/Renderer.html}. 
Notably, our rendering method, though not real-time, produces superior results taking an average of $ 12.86 (\pm 0.73)$ seconds to generate a 4k synthetic image.\footnote{With path tracing integrator using 800 samples to render each pixel, leveraging parallel computation of tiles on a cluster of 10 x RTX 2080Ti.}

\vspace{4pt}
\noindent\textbf{Balancing Gender and Racial Diversity:}
In response to the emerging concern of the under-representation of minority groups in the training datasets of recent medical AI solutions, our released dataset is specifically designed to cover greater diversity of cases. This initiative aims to promote fairer wound analysis by providing a more inclusive and representative dataset.

\vspace{4pt}
\noindent\textbf{3D wound reconstruction:}
Quantitative results for 3D wound reconstruction in two selected samples are reported in Table~\ref{table:reconstruction_results}. In our experiment, COLMAP demonstrates superior surface accuracy, while the performance of the Neural rendering-based method is nearly comparable.

\vspace{4pt}
\noindent\textbf{2D wound segmentation:}
Table~\ref{table:2Dseg_results}, presents the performance of SegFormer~\cite{xie2021segformer} trained on DFUC2022~\cite{kendrick2022translating}, tested on the synthetic images produced by \SthreeDW's model. 
The model, having been trained on real 2D wound data, exhibits promising performance when applied to our synthetic data, validating the quality of the \SthreeDW  ~dataset.
However, the limitations of 2D wound segmentations arise from the constrained perspective during capture, potentially impacting accuracy and comprehensiveness as they fail to fully represent the complexity of 3D structures (e.g., as shown in the second row of Fig.~\ref{fig:visualisation_outputs}, only the middle panel of leg/shoulder represents a complete view of a wound without presenting details such as depth).
Therefore, it is advisable to adopt methods that leverage rich 3D information through 3D segmentation. One way to achieve this is through projecting 2D masks onto 3D mesh vertices based on the results of the initial 2D segmentation.







\begin{figure}[!t]
 \centering
 \includegraphics[width=0.7\linewidth]{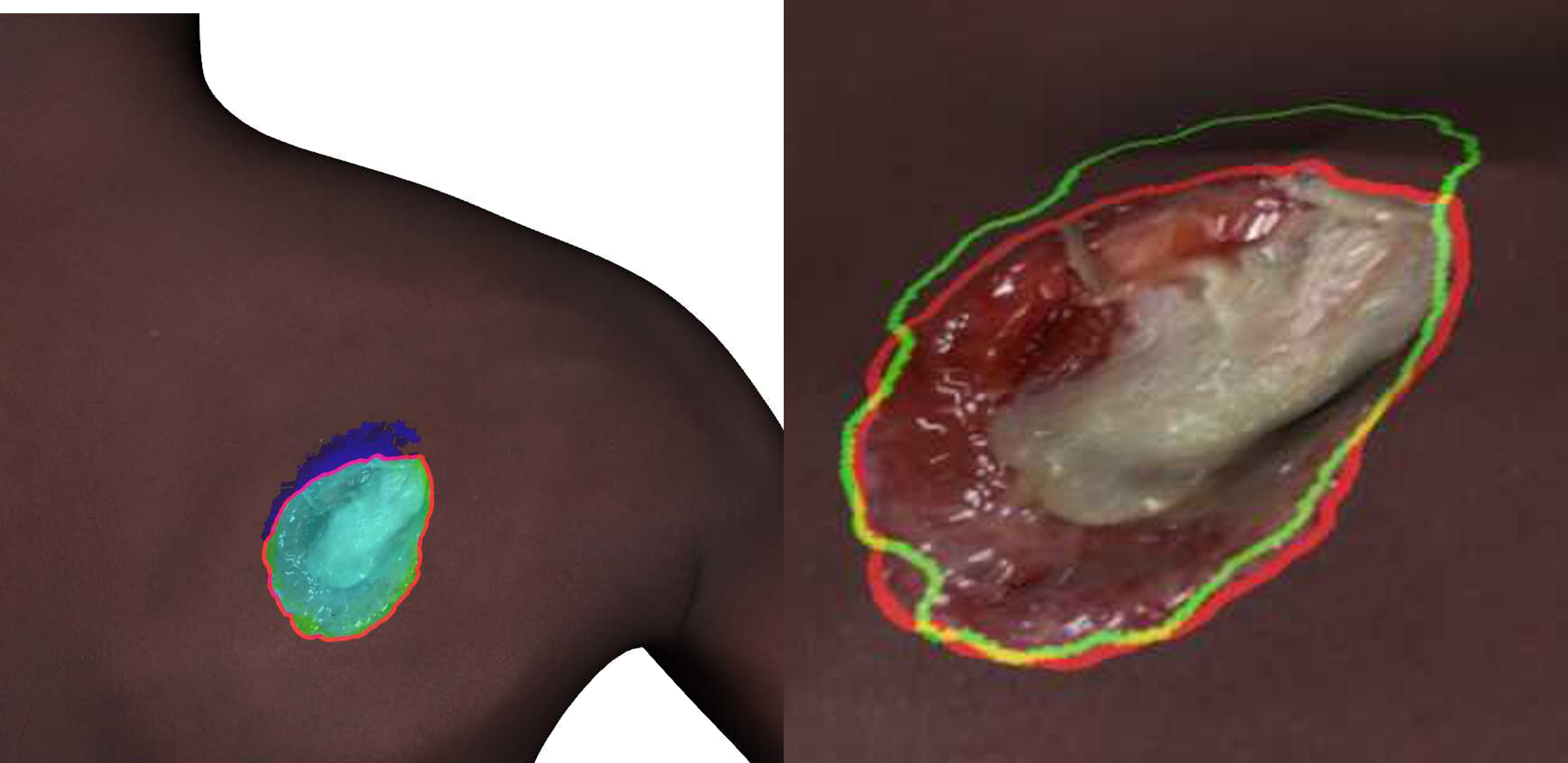}
 \vspace{-3pt}
 \caption{Color-coded 3D segmentation metrics for the shoulder wound.
 Right: ground-truth (red) and predicted (green) contour.
 }
 \label{fig:3D_segm1}
 \vspace{-6pt}
 \end{figure}

\vspace{4pt}
\noindent\textbf{3D wound segmentation:}
Table~\ref{table:3dwoundsegm_results} compares 3D wound segmentation results with ground truth using previously described metrics. Notably, for the second sample, incorporating a higher number of 2D segmentation maps enhances the performance of the resulting 3D segmentation.
Fig.~\ref{fig:3D_segm1} (left) shows the reconstructed 3D wound segmentation of the shoulder wound, generated from 120 renderings, with color-coded true positive (light blue), false positives (blue) and false negatives (yellow). 
The 3D projection of 2D segmentations provides a more precise understanding of the geometric failure modes of 2D segmentation models.


\vspace{-9pt}
\section{Conclusion}
In this paper, we contribute a unique 3D wound dataset to encourage collaboration between computer vision and medical imaging communities, intending to advance 3D wound reconstruction and documentation. We perform a study on widely used 3D reconstruction and segmentation pipelines, generating a set of baseline results pivotal for a better understanding of 3D wound analysis to address limitations in traditional 2D wound documentation.

\section{Compliance with Ethical Standards}
\vspace{-5pt}
This study was performed in line with the principles of the Declaration of Helsinki. The experimental procedures involving human subjects described in this paper were approved by CSIRO Health and Medical Human Research Ethics Committee (CHMHREC). The CHMHREC is an NHMRC Registered Human Research Ethics Committee (EC00187). CSIRO Ethics ID 2022\_025\_LR

\begin{table}[!t]
\caption{Evaluation of 2D segmentation model trained on DFUC2022~\cite{kendrick2022translating} and tested on renderings of the leg wound 3D model.}
\vspace{-6pt}
\centering
\label{table:2Dseg_results}
\resizebox{0.48\textwidth}{!}{%
\begin{tabular}{l l c c c}
\toprule
\textbf{Encoder} & \textbf{Network} & \textbf{Renderings} &  $\uparrow$ \textbf{IOU} & $\uparrow$ \textbf{Dice} \\
\midrule
MiT-B5              &  SegFormer~\cite{xie2021segformer} & 300 & 0.888  &  0.940  \\
\bottomrule
\multicolumn{5}{p{250pt}}
{
Mix Transformer encoders (MiT).
}
\end{tabular}}
\vspace{-6pt}
\end{table}

\begin{table}[!t]
\caption{Evaluation of 3D wound bed segmentation.}
\vspace{-6pt}
\centering
\label{table:3dwoundsegm_results}
\resizebox{0.48\textwidth}{!}{%
\begin{tabular}{l c c c}
\toprule
\textbf{Wound Sample} & $\downarrow$ \textbf{BAHD}~\textit{(mm)} & $\uparrow$ \textbf{P} & $\uparrow$ \textbf{R} \\
\midrule
Leg Wound, 300 renderings       & 0.028 &  0.925 & 0.985 \\
Shoulder Wound, 80 renderings   & 0.698 & 0.927 & 0.970 \\
Shoulder Wound, 120 renderings  & 0.101 & 0.957 & 0.971 \\
\bottomrule
\multicolumn{4}{p{300pt}}
{
BAHD: Balanced Average Hausdorff Distance; P: Precision; R: Recall. 
}
\end{tabular}}
\vspace{-6pt}
\end{table}

{
     \small
     \bibliographystyle{IEEEbib}
     \bibliography{ref}

\begin{thebibliography}{10}

\bibitem{MARTINENGO20198}
Laura Martinengo, Maja Olsson, Ram Bajpai, Michael Soljak, Zee Upton, Artur
  Schmidtchen, Josip Car, and Krister Järbrink,
\newblock ``Prevalence of chronic wounds in the general population: systematic
  review and meta-analysis of observational studies,''
\newblock {\em Annals of Epidemiology}, vol. 29, pp. 8--15, 2019.

\bibitem{litjens2017survey}
Geert Litjens, Thijs Kooi, Babak~Ehteshami Bejnordi, Arnaud Arindra~Adiyoso
  Setio, Francesco Ciompi, Mohsen Ghafoorian, Jeroen~Awm Van Der~Laak, Bram
  Van~Ginneken, and Clara~I S{\'a}nchez,
\newblock ``A survey on deep learning in medical image analysis,''
\newblock {\em Medical image analysis}, vol. 42, pp. 60--88, 2017.

\bibitem{zhang2022survey}
Ruyi Zhang, Dingcheng Tian, Dechao Xu, Wei Qian, and Yudong Yao,
\newblock ``A survey of wound image analysis using deep learning:
  classification, detection, and segmentation,''
\newblock {\em IEEE Access}, vol. 10, pp. 79502--79515, 2022.

\bibitem{wang2020fully}
Chuanbo Wang, DM~Anisuzzaman, Victor Williamson, Mrinal~Kanti Dhar, Behrouz
  Rostami, Jeffrey Niezgoda, Sandeep Gopalakrishnan, and Zeyun Yu,
\newblock ``Fully automatic wound segmentation with deep convolutional neural
  networks,''
\newblock {\em Scientific reports}, vol. 10, no. 1, pp. 21897, 2020.

\bibitem{oota2023wsnet}
Subba~Reddy Oota, Vijay Rowtula, Shahid Mohammed, Minghsun Liu, and Manish
  Gupta,
\newblock ``Wsnet: Towards an effective method for wound image segmentation,''
\newblock in {\em WACV}, 2023, pp. 3234--3243.

\bibitem{kendrick2022translating}
Connah Kendrick, Bill Cassidy, Joseph~M Pappachan, Claire O'Shea, Cornelious~J
  Fernandez, Elias Chacko, Koshy Jacob, Neil~D Reeves, and Moi~Hoon Yap,
\newblock ``Translating clinical delineation of diabetic foot ulcers into
  machine interpretable segmentation,''
\newblock {\em arXiv preprint arXiv:2204.11618}, 2022.

\bibitem{oota2021healtech}
Subba~Reddy Oota, Vijay Rowtula, Shahid Mohammed, Jeffrey Galitz, Minghsun Liu,
  and Manish Gupta,
\newblock ``Healtech-a system for predicting patient hospitalization risk and
  wound progression in old patients,''
\newblock in {\em WACV}, 2021, pp. 2463--2472.

\bibitem{ahmedt2023monitoring}
David Ahmedt-Aristizabal, Chuong Nguyen, Lachlan Tychsen-Smith, Ashley Stacey,
  Shenghong Li, Joseph Pathikulangara, Lars Petersson, and Dadong Wang,
\newblock ``Monitoring of pigmented skin lesions using 3d whole body imaging,''
\newblock {\em Computer Methods and Programs in Biomedicine}, vol. 232, pp.
  107451, 2023.

\bibitem{mirikharaji2023survey}
Zahra Mirikharaji, Kumar Abhishek, Alceu Bissoto, Catarina Barata, Sandra
  Avila, Eduardo Valle, M~Emre Celebi, and Ghassan Hamarneh,
\newblock ``A survey on deep learning for skin lesion segmentation,''
\newblock {\em Medical Image Analysis}, p. 102863, 2023.

\bibitem{Liu2019WoundImages}
Chunhui Liu, Xingyu Fan, Zhizhi Guo, Zhongjun Mo, Eric~I.Chao Chang, and Yan
  Xu,
\newblock ``{Wound area measurement with 3D transformation and smartphone
  images},''
\newblock {\em BMC Bioinformatics}, vol. 20, no. 1, pp. 1--21, 12 2019.

\bibitem{MirzaalianDastjerdi2019MeasuringAlgorithms}
Houman Mirzaalian~Dastjerdi, Dominique T{\"{o}}pfer, Stefan~J. Rupitsch, and
  Andreas Maier,
\newblock ``{Measuring surface area of skin lesions with 2D and 3D
  algorithms},''
\newblock {\em International Journal of Biomedical Imaging}, vol. 2019, 2019.

\bibitem{Shirley2019AWounds}
Tim Shirley, Dmitri Presnov, and Andreas Kolb,
\newblock ``{A lightweight approach to 3D measurement of chronic wounds},''
\newblock {\em Journal of WSCG}, vol. 27, no. 1, pp. 67--74, 2019.

\bibitem{Barbosa2020AccurateMotion}
Fellipe~M.C. Barbosa, Bruno~M. Carvalho, and Rafael~B. Gomes,
\newblock ``{Accurate chronic wound area measurement using structure from
  motion},''
\newblock {\em CBMS}, vol. 2020-July, pp. 208--213, 7 2020.

\bibitem{Sanchez-Jimenez2022SfM-3DULC:Area}
David S{\'{a}}nchez-Jim{\'{e}}nez, Fernando~F. Buch{\'{o}}n-Moragues, Begoña
  Escutia-Mu{\~{n}}oz, and Rafael Botella-Estrada,
\newblock ``{SfM-3DULC: Reliability of a new 3D wound measurement procedure and
  its accuracy in projected area},''
\newblock {\em International Wound Journal}, vol. 19, no. 1, pp. 44--51, 1
  2022.

\bibitem{shamsolmoali2021image}
Pourya Shamsolmoali, Masoumeh Zareapoor, Eric Granger, Huiyu Zhou, Ruili Wang,
  M~Emre Celebi, and Jie Yang,
\newblock ``Image synthesis with adversarial networks: A comprehensive survey
  and case studies,''
\newblock {\em Information Fusion}, vol. 72, pp. 126--146, 2021.

\bibitem{dai2021burn}
Fei Dai, Dengyi Zhang, Kehua Su, and Ning Xin,
\newblock ``Burn images segmentation based on burn-gan,''
\newblock {\em Journal of Burn Care \& Research}, vol. 42, no. 4, pp. 755--762,
  2021.

\bibitem{sinha2023dermsynth3d}
Ashish Sinha, Jeremy Kawahara, Arezou Pakzad, Kumar Abhishek, Matthieu Ruthven,
  Enjie Ghorbel, Anis Kacem, Djamila Aouada, and Ghassan Hamarneh,
\newblock ``Dermsynth3d: Synthesis of in-the-wild annotated dermatology
  images,''
\newblock {\em arXiv preprint arXiv:2305.12621}, 2023.

\bibitem{Renderpeople_bundle_swimwear}
Renderpeople,
\newblock ``{Bundle Swimwear Rigged 002},''
  \url{https://renderpeople.com/3d-people/bundle-swimwear-rigged-002/}, 2020.

\bibitem{saint20183dbodytex}
Alexandre Saint, Eman Ahmed, Kseniya Cherenkova, Gleb Gusev, Djamila Aouada,
  Bjorn Ottersten, et~al.,
\newblock ``3dbodytex: Textured 3d body dataset,''
\newblock in {\em 3DV}. IEEE, 2018, pp. 495--504.

\bibitem{wang2021foot}
Chuanbo Wang, Amirreza Mahbod, Isabella Ellinger, Adrian Galdran, Sandeep
  Gopalakrishnan, Jeffrey Niezgoda, and Zeyun Yu,
\newblock ``Fuseg: The foot ulcer segmentation challenge,''
\newblock {\em arXiv preprint arXiv:2201.00414}, 2022.

\bibitem{thomas2020medetec}
Steve Thomas,
\newblock ``Medetec wound database,'' 2020.

\bibitem{krkecichwost2021chronic}
Micha{\l} Kr{\k{e}}cichwost, Joanna Czajkowska, Agata Wijata, Jan Juszczyk,
  Bart{\l}omiej Pyci{\'n}ski, Marta Biesok, Marcin Rudzki, Jakub Majewski,
  Jacek Kostecki, and Ewa Pietka,
\newblock ``Chronic wounds multimodal image database,''
\newblock {\em Computerized Medical Imaging and Graphics}, vol. 88, pp. 101844,
  2021.

\bibitem{Kumar2019AWounds}
Syamantak Kumar, Dhruv Jaglan, Nagarajan Ganapathy, and Thomas~M Deserno,
\newblock ``A comparison of open source libraries ready for 3d reconstruction
  of wounds,''
\newblock in {\em Medical Imaging 2019: Imaging Informatics for Healthcare,
  Research, and Applications}. SPIE, 2019, vol. 10954, pp. 69--76.

\bibitem{Yu2022SDFStudio}
Zehao Yu, Anpei Chen, Bozidar Antic, Songyou Peng, Apratim Bhattacharyya,
  Michael Niemeyer, Siyu Tang, Torsten Sattler, and Andreas Geiger,
\newblock ``Sdfstudio: A unified framework for surface reconstruction,'' 2022.

\bibitem{COLMAP}
Johannes~Lutz Sch\"{o}nberger and Jan-Michael Frahm,
\newblock ``{COLMAP: A general-purpose Structure-from-Motion (SfM) and
  Multi-View Stereo (MVS) pipeline.},'' .

\bibitem{Meshroom}
AliceVision,
\newblock ``{Meshroom: A 3D reconstruction software.},'' 2018.

\bibitem{golub2013matrix}
Gene~H Golub and Charles~F Van~Loan,
\newblock {\em Matrix computations},
\newblock JHU press, 2013.

\bibitem{xie2021segformer}
Enze Xie, Wenhai Wang, Zhiding Yu, Anima Anandkumar, Jose~M Alvarez, and Ping
  Luo,
\newblock ``Segformer: Simple and efficient design for semantic segmentation
  with transformers,''
\newblock {\em Advances in Neural Information Processing Systems}, vol. 34, pp.
  12077--12090, 2021.

\bibitem{griwodz2021alicevision}
Carsten Griwodz, Simone Gasparini, Lilian Calvet, Pierre Gurdjos, Fabien
  Castan, Benoit Maujean, Gregoire De~Lillo, and Yann Lanthony,
\newblock ``Alicevision meshroom: An open-source 3d reconstruction pipeline,''
\newblock in {\em Proc. ACM Multimed. Syst. Conf.}, 2021, pp. 241--247.

\bibitem{taha2015metrics}
Abdel~Aziz Taha and Allan Hanbury,
\newblock ``Metrics for evaluating {3D} medical image segmentation: analysis,
  selection, and tool,''
\newblock {\em BMC Medical Imaging}, vol. 15, no. 29, 2015.

\bibitem{aydin2021usage}
Orhun~Utku Aydin, Abdel~Aziz Taha, Adam Hilbert, Ahmed~A Khalil, Ivana
  Galinovic, Jochen~B Fiebach, Dietmar Frey, and Vince~Istvan Madai,
\newblock ``On the usage of average hausdorff distance for segmentation
  performance assessment: hidden error when used for ranking,''
\newblock {\em European radiology experimental}, vol. 5, pp. 1--7, 2021.

\end{thebibliography}
}

\end{document}